\documentclass[10pt,conference]{IEEEtran}

\IEEEoverridecommandlockouts

\usepackage{epsfig,rotating,setspace,latexsym,amsmath,epsf,amssymb,amsfonts,bm,theorem,cite,algorithm,graphicx,epsf,authblk,epstopdf,color,algpseudocode,bbm}

\usepackage{algorithm} 
\usepackage{algpseudocode} 

\usepackage{subfigure}
\usepackage{mathtools,blkarray}
\usepackage{caption}
\usepackage{nicematrix}
\usepackage{multirow} 
\usepackage{lscape} 

\usepackage{subcaption}

\newcounter{procedure}
\newcounter{algorithm saved}

\makeatletter

\makeatletter
\newenvironment{procedure}[1][htb]{%
    \renewcommand{\ALG@name}{Procedure}
    \setcounter{algorithm saved}{\value{algorithm}} 
    \setcounter{algorithm}{\value{procedure}}
    \begin{algorithm}[#1]%
    }{\end{algorithm}
    \setcounter{procedure}{\value{algorithm}}
    \setcounter{algorithm}{\value{algorithm saved}}
}
\makeatother
\allowdisplaybreaks

\normalsize

\ifCLASSINFOpdf
 
\else

\fi

\begin{document}

\title{RCC-PFL: Robust Client Clustering under Noisy Labels in Personalized Federated Learning}
\author{Abdulmoneam Ali$~~~~~~\qquad$Ahmed Arafa\\Department of Electrical and Computer Engineering\\ University of North Carolina at Charlotte, NC 28223\\
$\quad$\emph{aali28@charlotte.edu}$\qquad$\emph{aarafa@charlotte.edu}
\thanks{This work was supported by the U.S. National Science Foundation under Grants CNS 21-14537 and ECCS 21-46099.}}

\markboth{IEEE Transactions on Communications}%
{Submitted paper}

\maketitle

\begin{abstract}
We address the problem of cluster identity estimation in a personalized federated learning (PFL) setting in which users aim to learn different personal models. The backbone of effective learning in such a setting is to cluster users into groups whose objectives are similar. A typical approach in the literature is to achieve this by training users' data on different proposed personal models and assign them to groups based on which model achieves the lowest value of the users' loss functions. This process is to be done iteratively until group identities converge. A key challenge in such a setting arises when users have noisy labeled data, which may produce misleading values of their loss functions, and hence lead to ineffective clustering. To overcome this challenge, we propose a label-agnostic \textit{data similarity}-based clustering algorithm, coined \textit{RCC-PFL},  with three main advantages: the cluster identity estimation procedure is \textit{independent} from the training labels; it is a \textit{one-shot} clustering algorithm performed prior to the training; and it requires fewer communication rounds and less computation compared to iterative-based clustering methods. We validate our proposed algorithm using various models and datasets and show that it outperforms multiple baselines in terms of average accuracy and variance reduction.
\end{abstract}

\section{Introduction}

Federated Learning (FL) is a privacy-preserving framework that enables users to learn a model without sharing their datasets \cite{pmlr-v54-mcmahan17a}. However, heterogeneity of user data is considered one of its main challenges as it can degrade learning accuracy, and has thus received significant attention \cite{adaptive_opt, ali2023delay}. One approach to tackle data heterogeneity is to rethink the entire learning paradigm; instead of learning a universal model that fits all users' data points, a personalized model can be learned for each user \cite{towards_pfl}. Learning multiple models by clustering users based on their learning objectives is a widely adopted paradigm, known as Personalized Federated Learning (PFL). 

In PFL, users seeking the same model can collaborate and learn it together. Thus, the main step is to identify users whose objectives are similar, i.e., solving the cluster identity problem \cite{clustered_FL, soft_clustering}, {\it without} revealing their objectives. The most common approach to address this challenge is based on the behavior of each user's training samples with respect to each personal model. Specifically, users are grouped based on which model minimizes their loss function \cite{clustered_FL, over_air_cfl}. The core idea is that, if the training samples of a pair of users behave similarly under a given model, then they may be interested in the same objective. However, this approach implicitly assumes that the loss function is being fed {\it clean} training samples. In practice, users may have {\it noisy} labeled data samples obtained, e.g., from low quality crowd-sourcing platforms, since high quality data is generally difficult to obtain \cite{noisy_label_sources}. The noisy label problem is more pronounced in FL because client-side data may be collected from diverse sources, each with its own stochastic noise. Thus, assuming high-quality labels for training data is overly optimistic, and relying on the loss function to solve the cluster identity problem needs to be reconsidered.

\textbf{Related Work.} To address the problem of noisy labels in machine learning, existing approaches can be divided into three main categories: detecting samples with noisy labels; refining the loss function; or following different training strategies. Detecting noisy-labeled samples typically centers around estimating the stochastic transition matrix between clean and noisy labels \cite{trans_estimation}, \cite{li2022estimating}. To avoid the challenge of estimating the transition matrix, the authors in \cite{yue2024ctrl} identify noisy labeled samples by observing the loss trajectory of the training of each sample. Based on this observation, each sample is classified as having a clean or noisy label. Refinement of the loss function aims to avoid overfitting by adding an extra regularization term to account for the presence of noise \cite{patrini2017making, wei2023mitigating}. The third approach involves training an auxiliary network for sample weighting or learning supervision \cite{jiang2018mentornet}, \cite{han2018coteaching}. In the FL setting, the authors in \cite{fang2022robust} follow a re-weighting approach that reduces the effect of the presence of noisy clients by reducing their weights' contribution when evaluating the loss function.

\textbf{Contributions.} Different from current approaches in the literature, we address the following question:
\begin{center}
\textit{Can we solve the cluster identity estimation problem for PFL without relying on the loss function or estimating the noise model for each user?}
\end{center} 
We answer the above question in the affirmative. In our proposed solution, we do not aim to detect noisy training samples or propose new noise-robust training algorithms. Instead, we focus on studying the
behavior of clustering users to learn multiple models in settings with significant label noise. We show that even without explicit data cleaning or noise-robust algorithms, a {\it one-shot data similarity-based clustering} algorithm can learn from data corrupted by arbitrary amounts of label noise. The key idea is that the proposed data similarity algorithm is label-agnostic, relying on features rather than labels. Therefore, the cluster identity estimation problem can be effectively resolved. An overview is depicted in Fig.~\ref{fig:clustering}.  

\noindent \textbf{Notation.} We use lowercase letters for scalars, bold lowercase letters for vectors, and bold uppercase letters for matrices throughout this paper. $\bm{A}[i,j]$ denotes the element in row $i$ and column $j$ of a matrix $\bm{A}$, and $\bm{A}[i]$ denotes its $i$th row. We represent the set $\{1,2,\dots,M\}$ by $[M]$.

\section{System Model}\label{Sys_Model}

We consider a federated learning system that has a parameter server (PS) and a set of users $\mathcal{K}=\{1,2,\dots,K\}$. Each user aims to learn a specific task from a total of $M$ different tasks. We define the tasks over the same dataset. In other words, given a dataset with $\{1,\dots, C\}$ class labels, each task is a subset of these labels, and the tasks are disjoint. That is, the $m$th task $T_m \subseteq \{1,\dots, C\}$, $\forall m \in [M]$, with $T_m \cap T_{\Bar{m}}= \phi$, $\forall m \neq \Bar{m}$. Since each user is interested in learning a specific task $T_{m}$, our system's objective is not to learn a single model, but rather to learn a number of personal models, equal to the number of tasks, in order to meet the users' requirements. We define the set of all tasks as $\mathcal{T}\triangleq\{T_1,\dots,T_M\}$.

Each user $k \in \mathcal{K}$ has a local data set $D_{k}=\{(x_{(k,i)},y_{(k,i)})\}_{i=1}^{n_k}$ where $x_{(k,i)}\in\mathbb{R}^p$ is the $i$th feature vector, $y_{(k,i)} \in \{1,\dots, C\}$ is the $i$th label, and $n_k$ is the number of training samples it has. We assume that most of a user's training samples belong to its intended learning task, denoted $\tau_k\in\mathcal{T}$, with some minority data from other unintended tasks $\mathcal{T}\setminus\tau_k$. In addition, we assume that all users are honest and trustworthy. Since different users may be interested in learning the same task, it is reasonable to group them together.

\begin{figure}[t]
    \centering
    \includegraphics[width=1\linewidth]{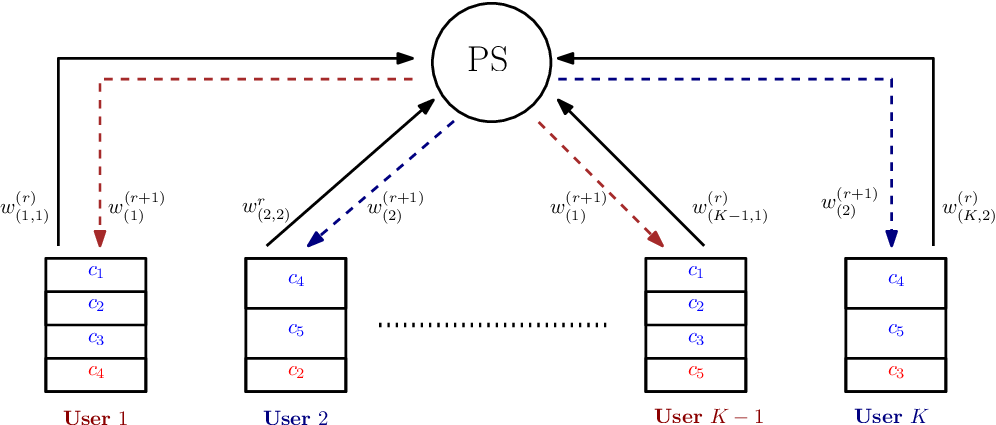}
    \caption{Illustration of our system model with two personal models: one model for learning $\{c_1,c_2,c_3\}$ and the other for learning $\{c_4,c_5\}$. Blue (resp. red) color represents clean (resp. noisy) labels. Here, users $1$ and $K-1$ should be clustered together. Same for users $2$ and $K$.}
    \label{fig:clustering}
    \vspace{-.2in}
\end{figure}

We assume that users' samples are not entirely clean, and that some of the labels are noisy. To model the unknown stochasticity of these noisy samples, we assume that each user $k$ has a portion $\alpha n_{k} $ of noisy labeled samples for some $\alpha\in(0,1)$. The main goal is to learn $M$ models by clustering users, without violating their privacy, ensuring that each cluster contains only users that aim to learn the same task. User $k$ belonging to cluster $m$ shares its model with the PS during the $r$th global communication round, denoted $w_{(k,m)}^{(r)}$, and the PS responds by sharing an aggregated model with all users in cluster $m$, denoted $w_{(m)}^{(r+1)}$, to initiate a new training round. An example setting is shown in Fig.~\ref{fig:clustering}.

Since we are interested in clustering users based on their intended learning tasks, we consider a worst-case noise scenario that hurts the clustering procedure the most. Specifically, we assume that the label-flipping procedure is {\it asymmetric} between labels in the sense that the true labels of a given task are all flipped into {\it one specific noisy label of another (unintended) task.} This increases confusion during the clustering procedure. We refer to this kind of noise as \textit{asymmetric task-dependent noise}, or \textit{task-flipping noise}. We note that this is a variation of the structure-biased noise described in \cite{rolnick1705deep}, in which clean labels are mapped to different (as opposed to only one in our case) noisy labels with different probabilities. 

Next, we illustrate in more detail how we perform the label-flipping procedure.

\section{Noise Models}

We focus on two noise models: {\it class-independent} noise and {\it class-dependent} noise. Both models share the same property that the clean labels to be flipped are mapped to one noisy label that is picked {\it uniformly} at random from other (unintended) tasks. However, they have some notable differences in how the clean labels are picked to be flipped, as we highlight next.

In the class-independent noise model, the training samples to be flipped are chosen {\it independently} of their true labels. Specifically, for user $k$, $\alpha n_k$ data points are sampled uniformly across all its training samples, and then they all get mapped to a label chosen uniformly at random from an unintended task $\mathcal{T}\setminus\tau_k$. However, in the class-dependent noise model, the training samples to be flipped are all chosen from a {\it particular} label from the intended task. Specifically, for user $k$, all $\alpha n_k$ data points are sampled from a specific label chosen randomly from $\tau_k$, and then are all mapped to one label from $\mathcal{T}\setminus\tau_k$. 

\begin{figure*}[t] 
    \centering
    \subfigure[Data with clean labels]{%
        \includegraphics[width=0.33\linewidth]{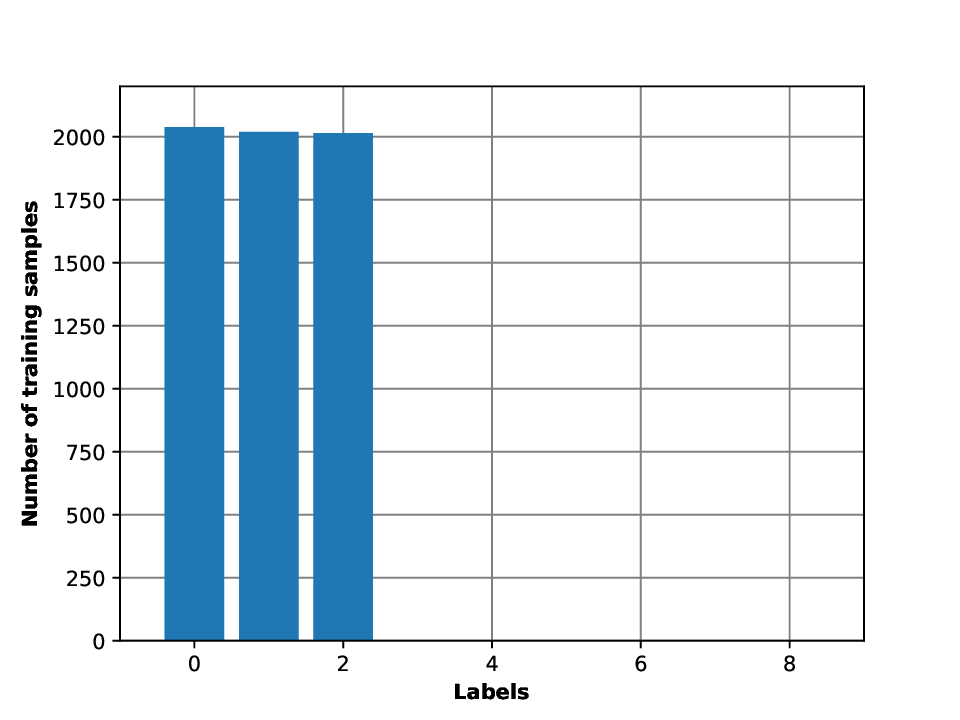}%
        \label{fig:a}%
        }%
    \hfill%
    \subfigure[Data with class-independent noisy labels]{%
        \includegraphics[width=0.33\linewidth]{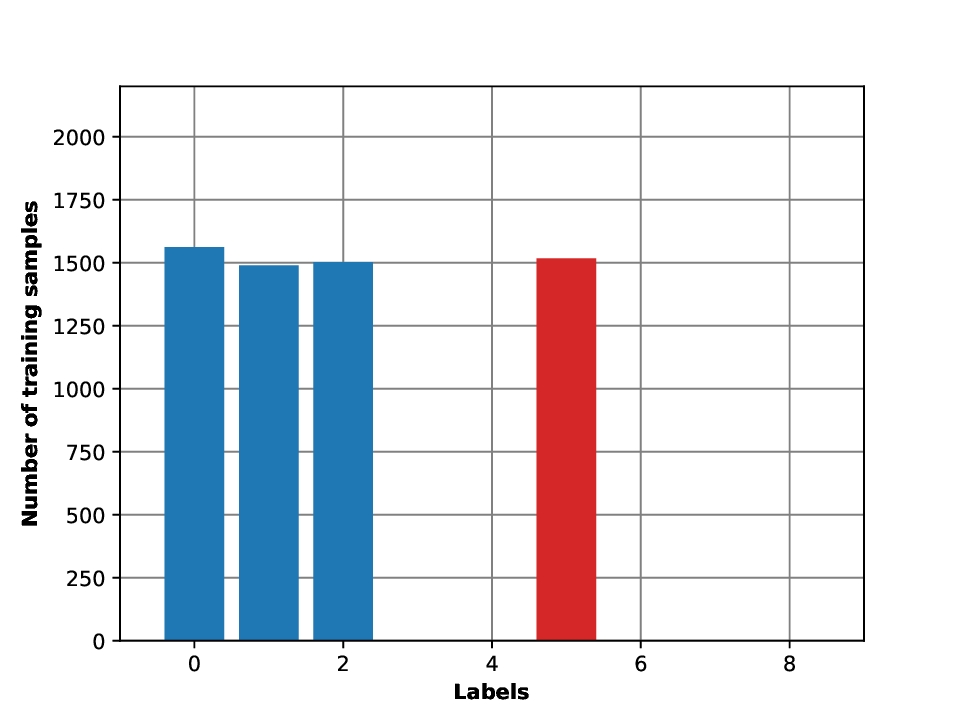}%
        \label{fig:b}%
        }%
        \hfill%
    \subfigure[Data with class-dependent noisy labels]{\includegraphics[width=0.33\linewidth]{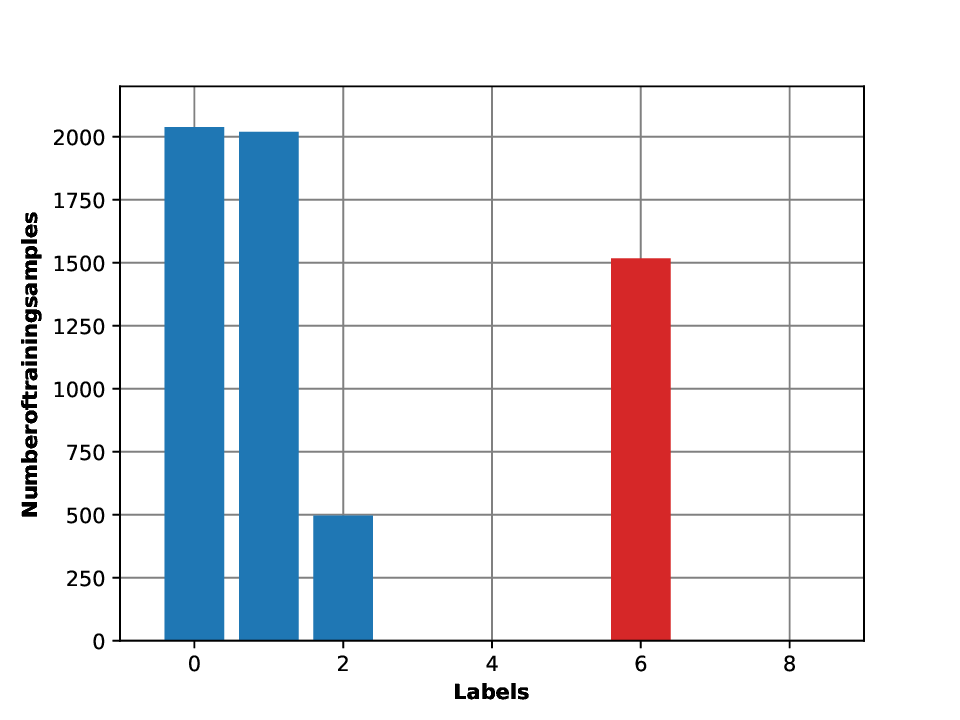}%
        \label{fig:c}%
        }
    \caption{Comparison between class-independent and class-dependent noise models on the Fashion-MNIST dataset with $\alpha=0.25$, and zero labels from unintended tasks. Blue color represents clean labels, and red represents noisy ones.}
    \label{fig:different_noisy_models}
    
\end{figure*}

Hence, the main difference between the two models is that in the class-independent model the drawn samples may have different labels, whereas in the class-dependent model, all samples are from a specific class label. We note that in the class-dependent model, if the chosen clean label's data are fewer than $\alpha n_k$, then the remaining ratio will be drawn from another label (also from $\tau_k$ if applicable). Consequently, under the class-dependent noise model, a user may end up with a fully corrupted label and one or more partially corrupted class labels. It is worth mentioning that sampling is performed without replacement, and thus, unlike the approach in \cite{rolnick1705deep}, there are no clean versions of noisy labeled samples. 

To better visualize the two noise models, Fig. \ref{fig:different_noisy_models} shows a realization of a user that has training samples from the Fashion-MNIST dataset. The intended task is to learn how to classify the labels $0$, $1$, and $2$. Fig. \ref{fig:a} shows the training samples with their ground truth labels, illustrating that, in this example, the user does not have any labels from its unintended tasks. Fig. \ref{fig:b} shows the results after applying the class-independent noise model. Setting $\alpha = 0.25$, we observe that approximately $\frac{1}{3}\times25\%$ of samples from each label are flipped to label $5$. Fig. \ref{fig:c}, on the other hand, shows the results after applying the class-dependent noise model. In this case, $25\%$ of the samples from label $2$ are flipped to label $6$.

The rationale and practicality of these two noise models stem from the fact that a user's data may be obtained from different sources, each with its own noise model. In the class-independent model, each data source contributes equally to the noisy labeled samples at the user. While in the class-dependent noise model, only one of the sources has specific corrupted label(s), while the other sources provide clean data.\footnote{In our setting, we follow one of the two noise models across all users.}

Next, we discuss our proposed clustering algorithm.

\section{Data Similarity PFL Clustering} \label{sec:data-valuation}

To enable users with the same personal model to learn cooperatively, they should be assigned the same cluster identity. Users seeking to learn the $m$th task should be grouped together to learn the model parameter $w_{m}$. Toward that end, each user $k$ in the $m$th cluster runs stochastic gradient descent (SGD) and shares its updated weight $w_{(k,m)}^{(r)}$ with the PS during global communication round $r$. The PS then uses FedAvg to send $w_m^{(r+1)}$ back to cluster $m$. The main challenge in applying the FedAvg algorithm among users with the same personal model is that their cluster identities are unknown. To efficiently estimate users'cluster identities, we adopt a data valuation-based technique that has been first proposed in \cite{data_valuation}, and then extended later in \cite{ds_asilomar24} in the context of hierarchical federated multi-task learning. The main idea is to measure the differences and similarities in the statistical properties of two datasets through the second moment. In this work, we extend the idea to demonstrate that the data similarity algorithm is a powerful technique, more robust in realistic and challenging scenarios, and outperforms different baselines. We also adopt a different method to extract informative features from raw data without relying on auxiliary information (cf. Section~\ref{sec:feature_mapping function}). 
 
 The data similarity algorithm consists of three main steps. The first step is that each user $i$ computes the eigenvalues and corresponding eigenvectors of their local dataset as follows:
\begin{align}\label{eq:eigens}
\bm{\lambda}_i,\bm{V}_i=\text{eigen}(\frac{1}{n_i}\Phi(\bm{X}_i)^{T}\Phi(\bm{X}_i)),
 \end{align}
 where  matrix $\bm{X}_i \in \mathbb{R}^{n_i\times p}$ denotes the arrangement of user $i$'s raw data, and $\Phi(\cdot)$ is a feature mapping function belonging to $\mathbb{R}^{n_i\times d}$ with $d < p$. The eigenvalues are stacked in a column vector $\bm{\lambda}_i \in \mathbb{R}^d$ and the associated eigenvectors are stacked in a matrix $\bm{V}_i \in \mathbb{R}^{d \times d}$.
The second step is that users exchange the eigenvector matrix to estimate eigenvalues.\footnote{Note that such exchange protects the privacy of the raw data, since the true eigenvalues are not shared.} The estimated eigenvalues are obtained by projecting the eigenvectors of other users onto their data and evaluating the Euclidean norm: 
\begin{align}
    \hat{\lambda}_{k}^{(j)}=\left\|\frac{1}{n_i}\Phi(X_i)^{T}\Phi(X_i) \bm{v}_{k}^{(j)}\right\|, ~\forall k \in [d], 
\end{align}
where $\bm{v}_{k}^{(j)}$ is the $k$th eigenvector of user $j$.
Based on the true eigenvalues and the estimated ones, each user can compute a metric called relevance/similarity as follows:
\begin{align}
   \lambda_{k}^{(i,j)}=& \frac{\min\{\lambda_k^{(i)},\hat{\lambda}_k^{(j)}\}} {\max\{\lambda_k^{(i)},\hat{\lambda}_k^{(j)}\}}, ~\forall k \in [d], \label{eq:lambd(i,j)}\\
    r(i,j)=&\prod_{k=1}^{d} (\lambda_{k}^{(i,j)})^{\frac{1}{d}},\label{eq:r(i,j)}
\end{align}
where the $\max\{.,.\}$ in \eqref{eq:lambd(i,j)} is used for normalization.
Finally, users share relevance values, computed in \eqref{eq:r(i,j)}, with the PS that in turn estimates the average data similarity between each pair of users as follows:
\begin{align}\label{eq:avg_r}
    \bm{R}(i,j)=\frac{r(i,j)+r(j,i)}{2}, ~ \forall i, j \in \mathcal{K}.
\end{align}
Note that $\bm{R}\in\mathbb{R}^{K\times K}$ is a symmetric matrix. The PS then runs the Hierarchical Agglomerative Clustering (HAC) algorithm \cite{pml1Book} to identify the cluster identity of each user. We arrange the cluster association of each user in a matrix $\bm{CI} \in \{0,1\}^{K\times M}$, where $\bm{CI}[k,m]=1$ indicates that user $k$ is associated with cluster $m$, and a user can associate with only one cluster. 
The detailed steps of the above are shown in Procedure \ref{proc:data_relevance}. 

Upon reaching this step, the PS can start training by broadcasting the initial weight $w_{m}^{(0)}$ for each cluster $m$. Each user $k$ in cluster $m$ runs SGD and shares its updated weight $w_{(k,m)}^{(1)}$ with the PS. The PS then aggregates the weights of each cluster separately, as shown in Step 9 of Algorithm \ref{alg:main_algo}, where $I_{(\bm{CI}[k]=m)}$ is an indicator function that equals one if user $k$ belongs to the cluster $m$. We coin the proposed algorithm RCC-PFL which is summarized
in Algorithm \ref{alg:main_algo}. 
\begin{algorithm}[t]
    \caption{RCC-PFL}\label{alg:main_algo}

\begin{algorithmic}[1]
\State \textbf{Input}: number of tasks $M$, number of users $K$, number of global iterations $G$, number of epochs $E$. 

\State\textbf{Cluster Identity}: $\bm{CI}~\leftarrow~$~Execute Procedure \ref{proc:data_relevance}.
\State \textbf{Training}: PS broadcasts $w_m^{(0)} ~\forall m\in [M]$
\For {$r \in [G]$}
\For{ $k \in [K]$}

\State $w_{k,m}^{(r)}\gets ~$ Perform SGD, $E$ iterations 
\EndFor
\State \textbf{PS aggregation}: 
\vspace{0.03in}
\State $w_m^{(r+1)}\!=\!\!\frac{1}{\sum_{k=1}^K \bm{CI}[k,m]}\sum_{k}I_{(\bm{CI}[k]=m)} w_{(k,m)}^{(r)},~ \forall m $

\EndFor
\end{algorithmic}
 \end{algorithm}
\begin{procedure}[t]
    
    \caption{Data Similarity Clustering}\label{proc:data_relevance}

\begin{algorithmic}[1]

        \State Perform an eigenvalue decomposition:
		\For {$i \in \mathcal{K}$}
				\State  $\bm{\lambda}_i,\bm{V}_i=\text{eigen}(\frac{1}{n_i}\Phi(\bm{X}_i)^{T}\Phi(\bm{X}_i))$
                \State Share $\bm{V}_i$ with the other users 
        \EndFor 
        \State Users compute the estimated eigenvalues:  
        \For {$i \in \mathcal{K}$}
            \For {$j \in \mathcal{K}$}
				\State  $\hat{\lambda}_{k}=\|\frac{1}{n_i}\Phi(X_i)^{T}\Phi(X_i) \bm{v}_{k}^{(j)}\|$,  $~\forall k \in [d]$
                \State Compute $\lambda_{k} ^{(i,j)}=\frac{\min\{ \lambda_k^{(i)}, \hat{\lambda}_{k}\}}{\max \{\lambda_k^{(i)}, \hat{\lambda}_{k}\}}$, $~\forall k \in [d]$
                \vspace{0.02in}
                \State Apply \eqref{eq:r(i,j)} and share with the PS
            \EndFor     

        \EndFor
		\State PS compute the average similarity matrix $\bm{R}$ using \eqref{eq:avg_r}
       \State Feed $\bm{R}$ and $M$ into the HAC algorithm and get $\bm{CI}$.
	\end{algorithmic}
 \end{procedure}

It is clear that our proposed algorithm for clustering users is based solely on features, as indicated by equation \eqref{eq:eigens}. Therefore, noisy labels do not affect the clustering decision. In contrast, most related work relies on the training process for clustering decision, which is significantly affected by noisy labels. Secondly, exchanging the matrix $\bm{V}$, which contains $d^2$ elements, is computationally less expensive than broadcasting the weights of each task to all users. Moreover, the matrix $\bm{V}$ is exchanged only once, whereas weight-based algorithms require repeated exchanges until the clustering decision converges. This is why we assert that the proposed RCC-PFL algorithm is more efficient, robust, and applicable in challenging settings.



\section{Experiments}\label{experiments}
We now present some experimental results evaluating the proposed RCC-PFL algorithm and demonstrating its robustness to noisy labels compared to multiple baselines. 

\textbf{Baselines.} We consider three baseline algorithms. The first is \textit{optimum} clustering, representing a genie-aided scheme that knows the users' tasks a priori. This serves as an upper bound.

The second baseline is an {\it iterative} clustering scheme denoted {\it IFCA-PFL}, a modified version of the scheme in \cite{clustered_FL} tailored to our PFL setting. It relies on the loss function to determine users' cluster identities. Specifically, we refine the original iterative algorithm in \cite{clustered_FL} (vanilla IFCA) by adding an additional step that enforces the required number of groups during each global communication round. The reason is that vanilla IFCA  does {\it not} address the possibility of users being merged into one cluster during the training rounds. We summarize IFCA-PFL in Algorithm~\ref{alg:iter_algo}. The algorithm consists of two main steps. The first step ensures that the system has the required number of clusters before starting the training. The second step, summarized in Procedure~\ref{proc:group_reassociation}, enforces the system to maintain the required number of tasks during each global communication round, and ensures that each user is associated with the model that minimizes their loss function. To reduce the computational complexity of IFCA-PFL, Procedure~\ref{proc:group_reassociation} may not be performed if the cluster identity of each user does not change after a number of consecutive rounds. 

The third baseline considered is a {\it single group} scheme, referring to the case where no clustering is performed, and all users learn a single global model (i.e., the vanilla FL setting). The purpose of including the single global scheme is to emphasize the necessity of clustering users. 

\begin{algorithm}[t]
    \caption{IFCA-PFL}\label{alg:iter_algo}

\begin{algorithmic}[1]
\State \textbf{Input}: number of tasks $M$, number of global iterations $G$, number of users $K$, number of epochs $E$.
\State \textbf{Cluster Identity}: $\bm{CI}~\leftarrow$ Execute Procedure~\ref{proc:group_reassociation}
\State PS broadcasts $w_{m}^{(0)}, m \in [M]$
\For {$r \in [G]$}
\For{ $k \in [K]$}
\State $w_{k,m}^{(r)}\leftarrow ~$ Perform SGD, $E$ iterations
\EndFor
\State \textbf{PS aggregation}: 
\vspace{0.03in}
\State $w_m^{(r+1)}\!\!=\!\!\frac{1}{\sum_{k=1}^K \bm{CI}[k,m]}\sum_{k}I_{(\bm{CI}[k]=m)} w_{(k,m)}^{(r)},~ \forall m$

\State \textbf{Cluster Re-association:} Execute Procedure \ref{proc:group_reassociation}
\EndFor

\end{algorithmic}
 \end{algorithm}

\begin{procedure} [t]
    \caption{Cluster Re-association}
    \label{proc:group_reassociation}

\begin{algorithmic}[1] 

\State Initialize $\bm{CI}$ as a $K\times M$ zero matrix
\For{ $k \in [K]$}
\State Estimate cluster identity: $\Hat{j}=\operatorname*{argmin}_{j \in [M]} F(w_{j}^{(r+1)}) $ 
\State Set $\bm{CI}[k,\Hat{j}]=1$ 
\EndFor\\
\textbf{PS Executes: }
\State counter = 0
\For{$m \in [M]$}
\If {$\sum_{k=1}^K\bm{CI}[k,m] \geq 1$}
\State counter = counter $+1$
\EndIf
\EndFor
\If{counter $= M$}
\State Return $\bm{CI}$
\ElsIf {Training has not started}
\State Reinitialize $M$ random $\{w_m^{(r+1)}\}$; go to Step 1
\Else
\State PS assigns to each cluster the users with the lowest loss function values evaluated at the cluster's model. 
\EndIf
\end{algorithmic}
\end {procedure}


 \textbf{Datasets and Models.}
We validate our proposed algorithm and baselines across different datasets, different learning models, and different number of tasks. 
We set the number of users $K=25$ and partition them equally among the number of tasks. Each task has a distinct set of classes. The task training samples are then distributed among the users of each task according to the Dirichlet distribution with parameter 2. In addition to the training samples belonging to the intended task, each user has a minority of training samples that belong to unintended tasks to test the robustness of our proposed algorithm. We assign 5\% of the labels from the other tasks to each user. We adopt 3 learning models and datasets: logistic regression with MNIST, two-layer perceptron with Fashion-MNIST, and CNN (as detailed in \cite{ds_asilomar24}) with CIFAR-10. For all experiments, we set the learning rate to $5\times 10^{-5}$, the number of epochs to 2, and the batch size to 32 for MNIST and Fashion-MNIST and 128 for CIFAR-10. To mitigate the impact of noisy labels during training, we fine-tune the weight decay (regularization) to be $0.001$. To accelerate training, the distributed Ray framework \cite{ray} is adopted. 

\textbf{Tasks.} We define the tasks as shown in Table \ref{tab:tasks}. The two tasks column has the following interpretation: learning odd versus even numbers for MNIST; learning clothes versus non-clothes images for Fashion-MNIST; and learning vehicles versus non-vehicle labels in CIFAR-10. As for the three and five tasks columns, we set a different combination of labels that reflects the diversity in users' preferences and also to demonstrate that the proposed algorithm does not depend on the task definition. 
\begin{table}[t]
 \begin{center}
    \caption{Tasks definition.}
    \label{tab:tasks}
   \resizebox{\columnwidth}{!} { \begin{tabular}{|c|c|c|c|}
    \hline
      \textbf{Dataset}  & \textbf{Two Tasks}&\textbf{Three Tasks} & \textbf{Five Tasks}\\
      \hline
      \multirow{3}{*}{MNIST} & $T_1=\{0,2,4,6,8\}$ &   $T_1=\{0,2,4\}$&   $T_1=\{0,2\},~ T_2=\{4,6\} $ \\ 
      & $T_2=\{1,3,5,7,9\}$&  $T_2=\{6,8,1\}$ & $T_3=\{8,1\},~ T_4=\{3,5\}$ \\ 
     & & $T_3=\{3,5,7,9\}$ & $T_5=\{7,9\}$ \\

      \hline

    \multirow{3}{*}{Fashion-MNIST} & $T_1=\{0,1,2,3,4,6\}$ &   $T_1=\{0,1,2\}$&   $T_1=\{0,1\},~ T_2=\{2,3\} $ \\ 
      & $T_2=\{5,7,8,9\}$&  $T_2=\{3,4,6\}$ & $T_3=\{4,6\},~ T_4=\{5,7\}$ \\ 
     & & $T_3=\{5,7,8,9\}$ & $T_5=\{8,9\}$ \\

      \hline
      
    \multirow{3}{*}{CIFAR-10} & $T_1=\{0,1,8,9\}$ &   $T_1=\{0,1,8\}$&   $T_1=\{0,1\},~ T_2=\{8,9\} $ \\ 
      & $T_2=\{2,3,4,5,6,7\}$&  $T_2=\{9,2,3\}$ & $T_3=\{2,3\},~ T_4=\{4,5\}$ \\ 
     & & $T_3=\{4,5,6,7\}$ & $T_5=\{6,7\}$ \\

      \hline
    \end{tabular}}
  \end{center}
\end{table}

\begin{table}[t]
 \begin{center}
    \caption{Results with class-independent noise.}
    \label{tab:class_indep}
   \resizebox{\columnwidth}{!} { \begin{tabular}{|c|c|c|c|c|}
    \hline
      \textbf{Dataset} & \textbf{Algorithms} & \textbf{Two Tasks}&\textbf{Three Tasks} & \textbf{Five Tasks}\\
      \hline
      \multirow{4}{*}{MNIST} & Optimum Clustering & 94.3 $\pm$ 0.6 &   96.2 $\pm$ 0.5 &   97.3 $\pm$ 0.3 \\ 
      & RCC-PFL & $\bm{94.4 \pm 0.5}$ &  $\bm{96.2 \pm 0.5}$ & $\bm{97.3 \pm 0.2}$ \\ 
      & IFCA-PFL  &  87.9  $\pm$ 1.9 & 91.4 $\pm$ 3.3 & 93.3 $\pm$ 4.7 \\
      & Single global model &  87.8 $\pm$ 1.0  & 86.9 $\pm$ 1.4 &84.9 $\pm$ 2.3\\
      \hline
      \multirow{4}{*}{Fashion-MNIST} & Optimum Clustering & 87.0 $\pm$ 0.3 & 92.0 $\pm$ 0.3 &95.6 $\pm$
0.2 \\ 
      & RCC-PFL & $\bm{87.0 \pm 0.3}$  & $\bm{92.0 \pm 0.3}$ & $\bm{95.6 \pm 0.3}$\\ 
      & IFCA-PFL  &  83.9 $\pm$ 2.1 & 90.4 $\pm$
2.3 & 92.5 $\pm$ 3.6 \\
      & Single global model &  83.9 $\pm $ 0.8 &79.7
$\pm$ 1.5& 79.2 $\pm$ 1.7\\
      \hline
    \multirow{4}{*}{CIFAR-10} & Optimum Clustering & 64.0 $\pm$ 0.6 & 75.4 $\pm$ 1.0& 85.8
$\pm$ 1.4\\ 
      & RCC-PFL & $\bm{63.9 \pm 0.8}$ & $\bm{75.5  \pm 1.0}$   & $\bm{85.8
\pm 1.3}$\\ 
      & IFCA-PFL  &  45.3 $\pm$ 4.4 &56.0 $\pm$ 12.7&72.7 $\pm$ 15.5 \\
      & Single global model &  46.7 $\pm$ 2.4 & 40.5 $\pm$ 3.7& 37.2 $\pm$ 10.4 \\
\hline
    \end{tabular}}
  \end{center}
\end{table}

\begin{table}[t]
 \begin{center}
    \caption{Results with class-dependent noise.}
    \label{tab:class_dep}
   \resizebox{\columnwidth}{!} { \begin{tabular}{|c|c|c|c|c|}
    \hline
      \textbf{Dataset} & \textbf{Algorithms} & \textbf{Two Tasks}&\textbf{Three Tasks} & \textbf{Five Tasks}\\
      \hline
      \multirow{4}{*}{MNIST} & Optimum Clustering & 89.1 $\pm$ 3.6 &   90.1 $\pm$ 4.1 &   96.4 $\pm$ 1.6\\ 
      & RCC-PFL& $\bm{89.2 \pm 3.5}$ &  $\bm{90.0 \pm 4.2}$ &  $\bm{96.4 \pm 1.6}$ \\ 
      & IFCA-PFL  &   84.9 $\pm$ 5.8 & 75.4 $\pm$ 10.8 &  88.2 $\pm$ 9.5 \\
      & Single global model &  81.0 $\pm$ 4.4  & 78.9 $\pm$ 3.5 & 82.0 $\pm$ 4.4 \\
      \hline
      \multirow{4}{*}{Fashion-MNIST} & Optimum Clustering & 83.3 $\pm$ 1.3 & 86.2 $\pm$ 6.4 &94.2 $\pm$
2.2 \\ 
      & RCC-PFL & $\bm{83.2 \pm 1.3}$  & $\bm{86.1 \pm 6.5}$ & $\bm{94.1 \pm 2.4}$\\ 
      & IFCA-PFL  & 79.1 $\pm$ 5.2 & 72.1 $\pm$
6.6 & 88.0 $\pm$ 8.1 \\
      & Single global model &  81.6 $\pm $ 1.2 & 76.1 
$\pm$ 4.6 & 79.2 $\pm$ 2.5 \\
      \hline
    \multirow{4}{*}{CIFAR-10} & Optimum Clustering & 57.8 $\pm$ 2.2 & 69.5 $\pm$ 2.9 &86.9 $\pm$ 1.2\\ 
      & RCC-PFL & $\bm{57.8 \pm 2.1}$  & $\bm{ 69.3 \pm 2.7}$ & $\bm{86.9 \pm 1.3}$\\ 
      & IFCA-PFL  &  42.4 $\pm$ 5.1 & 44.6 $\pm$ 6.8 & 60.3 $\pm$ 16.2\\
      & Single global model &  46.5 $\pm$ 2.6  & 42.6 $\pm$ 3.0 &38.05 $\pm$ 4.9 \\
\hline
    \end{tabular}}
  \end{center}
\end{table}

\begin{figure*}[htp] 
    \centering
    \subfigure[Group 1 Loss]{%
        \includegraphics[width=0.33\linewidth]{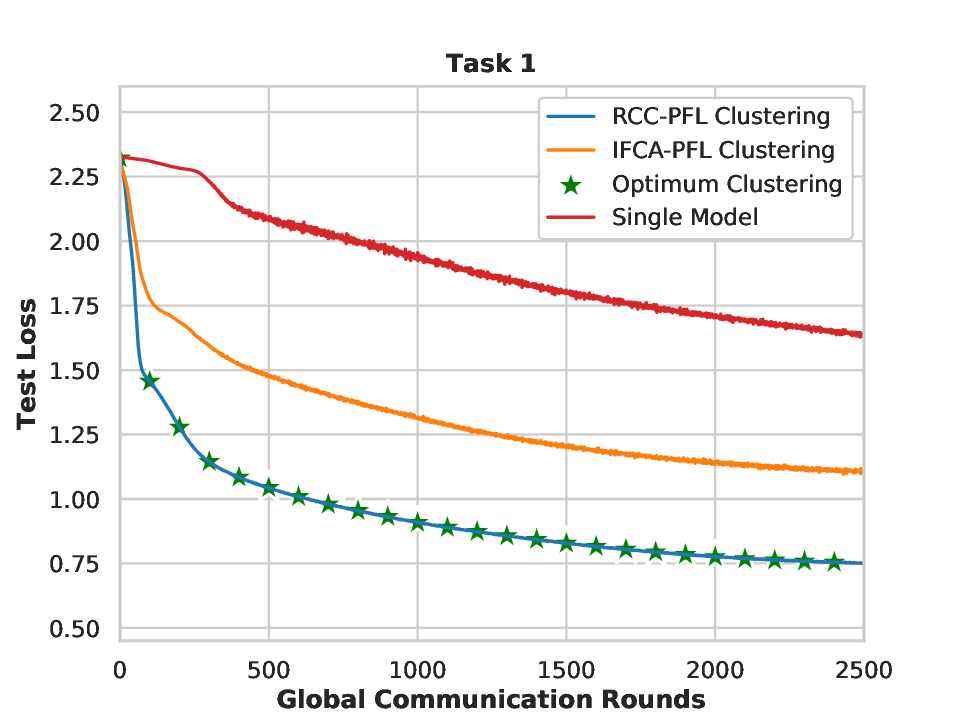}%
        \label{fig:loss_1}%
        }%
    \hfill%
    \subfigure[Group 2 Loss]{%
        \includegraphics[width=0.33\linewidth]{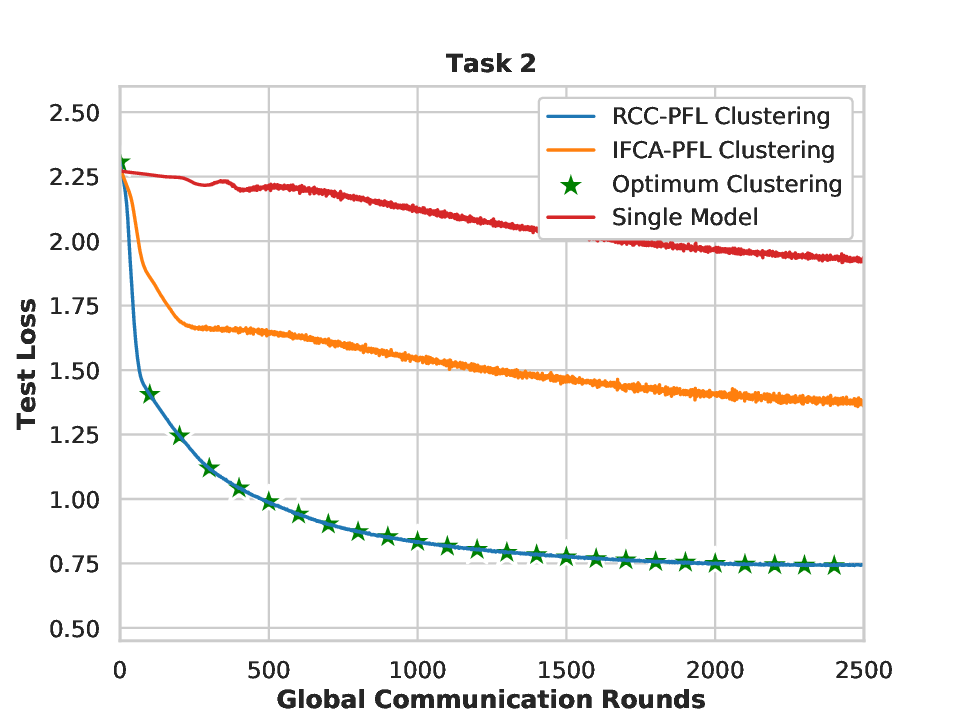}%
        \label{fig:loss_2}%
        }%
        \hfill%
    \subfigure[Group 3 Loss]{\includegraphics[width=0.33\linewidth]{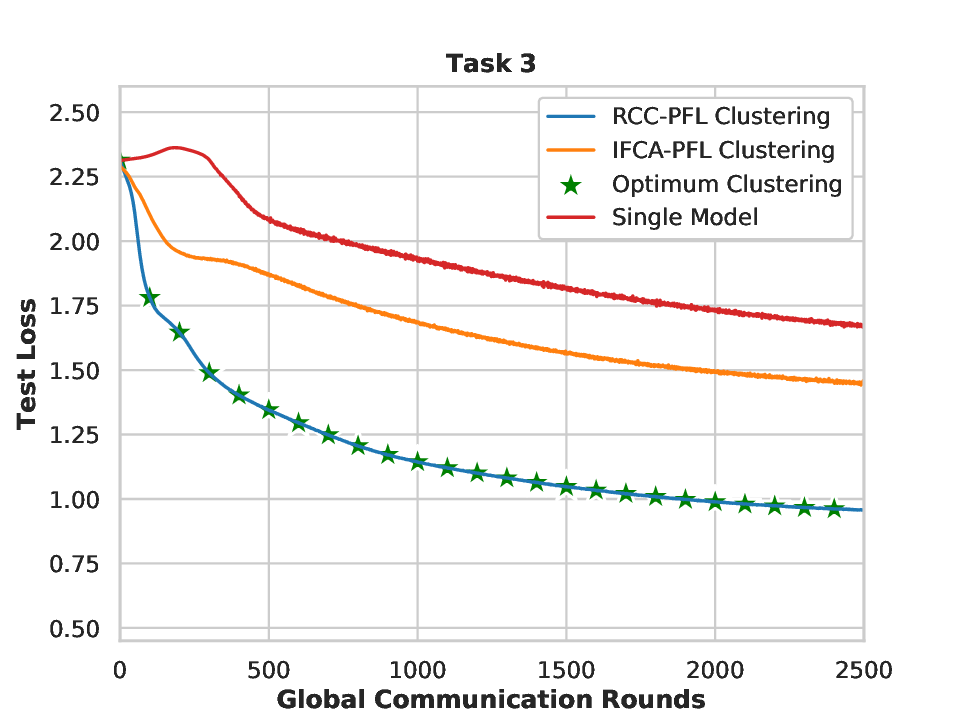}%
        \label{fig:loss_3}%
        }
    \caption{Comparison between three groups' performance on CIFAR-10 under class-independent noise.}
    \label{fig:loss_3_grps}
\end{figure*}

\textbf{Feature Mapping Function.}\label{sec:feature_mapping function}
There are two advantages of the feature mapping function $\Phi$: extracting informative features from the raw data, and reducing their dimensionality. There are various methods to design such a function, with the most common approach being based on a pre-trained model \cite{data_valuation}, \cite{Many-taskFL}. However, this approach requires users to have access to a well-performing model prior to training, which may not always be practical. Therefore, our work adopts another powerful feature extraction method that does not require additional information: the histogram of oriented gradients (HoG) \cite{HoG}. The main idea of the HoG is to divide the image into cells, compute the orientation and magnitude of the gradient for each cell, and then aggregate the gradient information into a histogram of oriented gradients. These histograms describe the features of the image and help detect objects within it. We note that HoG relies solely on the feature vector of an image, making it insensitive to noisy labels. 

\textbf{Beating the Baselines.} We conduct five experimental runs and
calculate the average test accuracy across these experiments.
 Tables \ref{tab:class_indep} and \ref{tab:class_dep} demonstrate the superiority of the RCC-PFL algorithm in perfectly clustering users and achieving the highest performance, while the iterative scheme, IFCA-PFL, fails to identify the correct clusters, resulting in deteriorated performance. The results also show that IFCA-PFL has a higher variance, which worsens with more challenging data and additional tasks. 
 
 The impact of the two noisy models on performance is evident in the average accuracy; performance under class-independent noise is better than that under class-dependent noise. Our justification is that the noise adversely affects the performance more when it is concentrated on certain label(s) (e.g., a user may have one or more labels that are completely corrupted and simultaneously lack a clean version of such labels). In contrast, in class-independent noise, the noise is distributed across different labels; all user's labels are affected, but the user still has a clean portion of each. In addition, collaborative learning between users can mitigate the effect of these noisy labeled samples. 
 
 It is also clear that learning different models is more important than using the universal one, and the PFL setting becomes vital with a greater variety of tasks. For example, in CIFAR-10, the more diverse the personal models are, the worse the performance of the universal model becomes.
Additionally, we observe that personalized learning in the presence of noisy labels becomes more feasible (higher accuracy) when having a small set of intended learning labels in each task. 

Finally, Fig. \ref{fig:loss_3_grps} shows the testing loss for each individual group/task on CIFAR-10. It is clear that each group follows a different loss trajectory, even though all groups begin training based on the same initial weights. This highlights that the learning behavior also depends on the task definition. The proposed RCC-PFL clustering algorithm successfully assigns the users' cluster identities, and hence its performance coincides with the optimum clustering. IFCA-PFL, however, fails to infer the correct cluster identities, and hence its loss value is larger.

\section{Conclusion}\label{conclusion}

In this paper, RCC-PFL, a one-shot data similarity-based algorithm that aims to solve the cluster identity estimation problem in a personalized federated learning setting under noisy labeled data, has been introduced. The superiority of RCC-PFL compared to multiple baselines has been shown in terms of efficiently estimating the cluster identities of the users and consequently achieving higher performance. Furthermore, our proposed algorithm requires only a single global iteration with the PS, resulting in significant communication cost savings. One direction for future investigations is to study the robustness of our proposed algorithm in the presence of malicious users.

\ifCLASSOPTIONcaptionsoff
  \newpage
\fi

\bibliographystyle{unsrt}
\bibliography{Ali}

\end{document}